\documentclass{article}

\usepackage{amssymb}
\usepackage{graphicx}
\usepackage{booktabs}

\usepackage[preprint]{neurips_2025}

\usepackage[utf8]{inputenc} 
\usepackage[T1]{fontenc}    
\usepackage[hidelinks]{hyperref}
\usepackage{url}            
\usepackage{booktabs}       
\usepackage{amsfonts}       
\usepackage{nicefrac}       
\usepackage{microtype}      
\usepackage{xcolor}         
\usepackage{amsmath}

\usepackage{comment}

\definecolor{anscolor}{rgb}{0, 0.016, 0.6}

\title{Density-Informed VAE (DiVAE): \\ Reliable Log-Prior Probability via Density Alignment Regularization}

\author{
Michele Alessi$^{1,2}$, Alessio Ansuini$^{2}$, Alex Rodr\'{\i}guez$^{1}$\\[0.8em]
$^{1}$Department of Mathematics, Informatics and Geosciences, University of Trieste, Italy\\
$^{2}$Laboratory of Data Engineering, Area Science Park, Trieste, Italy\\[0.8em]
\texttt{michele.alessi@phd.units.it}\\
\texttt{alessio.ansuini@areasciencepark.it}\\
\texttt{alejandro.rodriguezgarcia@units.it}
}

\begin{document}
\maketitle
\begin{abstract}
We introduce \textbf{Density-Informed VAE (DiVAE)}, a lightweight, data-driven regularizer that aligns the VAE log-prior probability $\log p_Z(z)$ with a log-density estimated from data. Standard VAEs match latents to a simple prior, overlooking density structure in the data-space. DiVAE encourages the encoder to allocate posterior mass in proportion to data-space density and, when the prior is learnable, nudges the prior toward high-density regions. This is realized by adding a robust, precision-weighted penalty to the ELBO, incurring negligible computational overhead. On synthetic datasets, DiVAE (i) improves distributional alignment of latent log-densities to its ground truth counterpart, (ii) improves prior coverage, and (iii) yields better OOD uncertainty calibration. On \textsc{MNIST}, DiVAE improves alignment of the prior with external estimates of the density, providing better interpretability, and improves OOD detection for learnable priors.
\end{abstract}

\section{Introduction}\label{Introduction}
Variational autoencoders (VAEs) typically assume a simple prior $p_Z$ (e.g., $\mathcal{N}(0,I)$) \cite{kingma2013auto}. This choice encourages the variational posterior $q_\phi(z\mid x)$ to place latent codes near the origin and to distribute mass according to the prior rather than the data’s \emph{empirical density}. As a result, the log-prior probability $s(z)=\log p_Z(z)$ may be poorly aligned with how samples concentrate in data space, reducing its usefulness for anomaly detection, out-of-distribution (OOD) scoring, and uncertainty estimation. It also contributes to well-known issues such as the \emph{hole problem} --regions of high prior mass unsupported by the theoretically optimal prior: the aggregated variational posterior \cite{tomczak2024deep,rezende2018tamingvaes}.

To address the limitations of a simple, fixed prior described above, we introduce \textbf{Density-Informed VAE (DiVAE)}, which injects a per-sample density signal $\rho$ --an estimate of log-density measured in a data-derived projection-- into training via a regularization term. 
Acting as a \emph{density teacher}, DiVAE adds an alignment penalty that pushes the encoder to place latent codes in proportion to $\rho$. Consequently, even with a non-learnable prior (e.g., $\mathcal{N}(0,I)$), the latent space recovers the data-space density structure, improving the usefulness of $s(z)$ without changing the prior itself.
For a learnable prior (e.g., a Gaussian mixture model, GMM), the same density signal simultaneously (i) guides the encoder toward regions whose mass reflects the estimate and (ii) pulls the prior parameters toward those high-density regions, thereby improving agreement between prior mass and empirical structure. 
Concretely, we augment the ELBO with a robust, precision-weighted alignment loss (Section~\ref{Methods}) that incurs negligible computational overhead (Table~\ref{tab:mnist_timing}) and requires no changes to the encoder or decoder architectures.

We implement this alignment in two forms: a \emph{direct} variant that matches the model’s latent log-prior to the external log-density estimate, and a \emph{flow-corrected} variant that learns a normalizing flow to account for the change-of-variables (Jacobian) between latent and data-derived projections.

We show that on synthetic datasets (in which we know the ground truth), DiVAE (Sec.~\ref{Results}): 
(i) improves distributional alignment of latent log-densities to the ground truth log-density,  shifting the mean toward the true scale;
(ii) improves prior coverage, and 
(iii) yields better OOD uncertainty calibration—stronger separation, higher posterior entropy. 
These gains hold under both standard-normal and learnable GMM priors; \textsc{flow} attains the strongest separation, whereas \textsc{direct} provides the best calibration trade-off.
In MNIST, where the ground truth is unknown, we find stable improvements in alignment between several types of prior and an external estimate of the density, with latent codes that are significantly most typical under the learned prior (see Table \ref{tab:mnist_result}); for the OOD, when the prior is learnable, there are consistent improvements in adopting a density regularizer (\ref{tab:mnist_ood}) .

\paragraph{Related Work}\label{Related}
A major line of work improves VAE priors to better approximate the (theoretically optimal) aggregated posterior. Building on the observation that the aggregate is typically a gaussian mixture \cite{tomczak2024deep}, VampPrior introduces learnable pseudo-inputs so that the prior becomes a mixture of variational posteriors that explicitly targets this distribution \cite{tomczak2018vae}. Flow– and autoregressive–based priors further increase flexibility and improve likelihoods/compression 
\cite{rezende2015variational,kingma2016iaf,chen2016variational,gulrajani2017pixelvae,papamakarios2021normalizing}.

In this work, rather than only enriching the prior, we directly align the density around data points --measured in the PCA-reduced space-- with the latent density, enforcing local agreement between empirical structure and prior mass.

\section{Methods}\label{Methods}
\paragraph{Setup}
Let $x\in\mathcal{X}$ be an input,  $q_\phi(z\mid x)=\mathcal{N}\!\big(\mu_\phi(x),\mathrm{diag}(\sigma_\phi^2(x))\big)$ is the encoder, $p_\theta(x\mid z)$ the decoder, and $p_Z(z) = p_Z(z; \psi)$ a prior (fixed or learnable) with parameters $\psi$ (for a standard Gaussian prior, $\psi=\varnothing$). 

We apply a dimensionality-reduction step used \emph{only} for density estimation and for comparing densities between data and latent spaces. Specifically, we project the data from the original space $\mathcal{X}$ onto the linear subspace $\mathcal{U}\subseteq\mathcal{X}$ spanned by the top $d$ principal components, where $d=\dim(\mathcal{Z})$ matches the latent dimensionality. Let $P:\mathcal{X}\!\to\!\mathcal{U}$ denote the PCA projector; for each training point $x_i$, we write $u_i=P\,x_i$ for its projection.

We assume that we have access to a \emph{external} (i.e., model-independent) log-density estimate $\rho_i$ in the $\mathcal{U}$ space of the PCA projected data, with a 
standard error $\sigma_i$ expressing the estimator’s uncertainty at $u_i$. In practice, this estimate will be performed with standard techniques such as Kernel-Density Estimation (KDE), or Point Adaptive k-Nearest Neighbors (PAk) \cite{silverman2018density,PAk2018}.

We define the \emph{log-prior probability} for $x_i$, expressing the log probability assigned by the prior to the encoding of a data point $x_i$, as $s_i = \mathbb{E}_{z\sim q_\phi(\cdot\mid x_i)}\!\big[\log p_Z(z)\big]$, and estimate it with Monte Carlo sampling, where in practice we use one-sample estimates. 
The central idea of our work is to nudge the log-prior probabilities $\left \{s_i \right \}$ towards $\left \{\rho_i \right \}$ using a penalty, which we call the \emph{alignment loss}. 
This loss comes in two flavours, depending on the aligner type: 1) a direct aligner and 2) a flow-corrected aligner.  

\paragraph{Direct aligner}

The direct aligner operates with a robust, precision-weighted loss term that acts as a regularizer, and compares the model’s prior log-densities $\left \{s_i \right \}$ to the external log-densities $\left \{\rho_i \right \}$.
We define precision weights from the estimator’s uncertainties $w_i =\sigma_i^{-2}.$ 

Given a mini–batch $\{x_i\}_{i=1}^B$ the direct alignment loss is written as: 

\begin{equation}
\label{eq:aligner-batch}
\mathcal{L}_{\text{direct-align}} =
\sqrt{\frac{1}{B}\sum_{i=1}^B w_i\, \Gamma_\delta\!\big(s_i -\rho_i\big)},
\end{equation}
where $\Gamma_\delta$ is the Huber loss. The Huber loss \cite{huber1992robust} behaves quadratically within a range $\delta>0$, and linearly outside this range, to mitigate the influence of outliers (see \ref{Model_details}). In our experiments we set $\delta = 1$, corresponding to $\approx \frac{1}{2}$ probability for the residual to fall in the linear regime. 

\paragraph{Training objective} We optimize the (negative) ELBO plus the alignment term $\mathcal{L} = -\,\mathcal{L}_{\text{ELBO}} + \gamma_t\,\mathcal{L}_{\text{direct-align}}$, with a warm-up weight $\gamma_t$, practically chosen as a linear warm-up from $0$ to a $1$ over training epochs. Unless stated otherwise, we \emph{do not} detach encoder gradients in \eqref{eq:aligner-batch}; this allows the regularizer to shape the variational posterior $q_\phi$.

\paragraph{Flow-corrected aligner}

The direct aligner supervises the log-prior probability $s(z)=\log p_Z(z)$ but, by itself, cannot account for the change of variables between the latent space $\mathcal{Z}$ and $\mathcal{U}$, i.e., it omits the log-determinant Jacobian term. We therefore learn an invertible map $f:\mathcal Z\!\to\!\mathcal U$, parameterized as a normalizing flow \cite{papamakarios2021normalizing}. The change of variables formula implies $\log p_U(u)= \log p_Z\!\big(f^{-1}(u)\big) + \log\big|\det J_{f^{-1}}(u)\big|.$
We train $f$ by maximum likelihood on observed projections $u_i$:
\begin{equation}
\label{eq:flow-mle}
\mathcal{L}_{\text{flow-ML}}
= -\frac{1}{B}\sum_{i=1}^B \Big[\log p_Z\!\big(f^{-1}(u_i)\big) + \log\big|\det J_{f^{-1}}(u_i)\big|\Big].
\end{equation}
For alignment in $\mathcal Z$-space, the ideal target would be
$s(z)\;=\; \log p_U\!\big(f(z)\big) + \log\big|\det J_f(z)\big|$.
Since $p_U$ is unknown, we substitute the external density estimate at the corresponding projected point. Assuming local coherence $f(z_i)\approx u_i$, we obtain the practical approximation $s_i\;\approx\; \rho_i \;+\; \log\big|\det J_f(z_i)\big|$.
The flow-corrected alignment loss then reads
\begin{equation}
\label{eq:aligner-flow}
\mathcal{L}_{\text{flow-align}}
\;=\;
\sqrt{\frac{1}{B}\sum_{i=1}^B w_i\, \Gamma_\delta\!\Big(s_i - \rho_i - \underbrace{\log\big|\det J_f(z_i)\big|}_{\text{stop-grad w.r.t.\ }f}\Big)}
\end{equation}
where the log-determinant term provides the correct Jacobian correction, and we stop its gradient into $f$ during alignment (so the encoder/prior is driven by the density signal while the flow is learned by \eqref{eq:flow-mle}). 
The overall objective becomes $\mathcal{L}_{\text{total}}= -\mathcal{L}_{\text{ELBO}}+\gamma_t\,\mathcal{L}_{\text{flow-align}} +\mathcal{L}_{\text{flow-ML}}$ optimized with two parameter blocks:   $(\phi,\psi,\theta)$ for the VAE and flow parameters for $f$.  


\begin{table}
\centering
\begin{tabular}{llrrrrrrrr}
\toprule
& & \multicolumn{2}{c}{ $k{=}4$} & \multicolumn{2}{c}{ $k{=}8$} & \multicolumn{2}{c}{ $k{=}12$} \\
\cmidrule(lr){3-4}\cmidrule(lr){5-6}\cmidrule(lr){7-8}
Prior & Method & KS$\downarrow$ & KL$\downarrow$ & KS$\downarrow$ & KL$\downarrow$ & KS$\downarrow$ & KL$\downarrow$ \\
\midrule
GMM      & \textsc{none}   & 0.469       & 114.184       & 0.501       & 20.877        & 0.380       &   \textbf{3.728}     \\
GMM      & \textsc{direct} & \textbf{0.102}       & 69.620        & \textbf{0.073}       & 6.447         & \textbf{0.124}       &   6.823     \\
GMM      & \textsc{flow}   & 0.946       & \textbf{4.419}         & 0.753       & \textbf{5.334}         & 0.934       &   4.582     \\
\midrule
STANDARD & \textsc{none}   & \textbf{0.116}       & -             & 0.322       & -             & 0.418       &   -     \\
STANDARD & \textsc{direct} & 0.228       & -             & \textbf{0.231}       & -             & 0.384       &   -     \\
STANDARD & \textsc{flow}   & 0.338       & -             & 0.315       & -             & \textbf{0.255}       &   -    \\
\bottomrule
\end{tabular}
\caption{KS statistic and coverage divergence
$\mathrm{KL}(p_2\,\|\,p_Z(\cdot;\psi))$ on synthetic Gaussian-mixture
datasets with $k{=}4,8,12$ components. Means over three seeds. External
densities are estimated using PAk. KL is omitted for the standard-normal
prior since $p_Z$ is fixed and identical across methods.}
\label{tab:ks-kl-main}
\end{table}

\begin{table}
    \centering
    \begin{tabular}{llcccc}
        \toprule
        Prior & Method & $\overline \ell \uparrow$ & $\bar{s} \uparrow$ & KS $\downarrow$ & W $\downarrow$  \\
        \midrule
        STANDARD & \textsc{none}   &\textbf{-101.214 $\pm$ 29.087} &   -14.209 $\pm$ 1.750 & \textbf{0.1102} & 0.8374  \\
        STANDARD & \textsc{direct} & -101.517 $\pm$ 28.980 &   \textbf{-13.958 $\pm$ 1.629} & 0.1133 & \textbf{0.6475}  \\
        STANDARD & \textsc{flow}   & -101.585 $\pm$ 28.660 &   -16.102 $\pm$ 2.053 & 0.4540 & 2.7300  \\
        \midrule
        GMM      & \textsc{none}   & -100.087 $\pm$ 29.314 &   -20.633 $\pm$ 2.510 & 0.9008 & 7.2613 \\
        GMM      & \textsc{direct} & \textbf{-99.793 $\pm$ 29.419} &   \textbf{-13.568 $\pm$ 3.216} & \textbf{0.0815} & \textbf{0.3992}\\
        GMM      & \textsc{flow}   & -101.309 $\pm$ 29.402 &   -41.539 $\pm$ 2.024 & 1.0000 & 28.1673  \\
        \midrule
        VAMP      & \textsc{none}   & \textbf{-99.548 $\pm$ 29.393} &   -20.833 $\pm$ 2.875 & 0.8873 & 7.4612 \\
        VAMP      & \textsc{direct} & -99.895 $\pm$ 29.503 &   \textbf{-13.510 $\pm$ 3.167} & \textbf{0.0698} & \textbf{0.3460} \\
        VAMP      & \textsc{flow}   & -102.175 $\pm$ 30.056 &   -56.126 $\pm$ 1.757 & 1.0000 & 42.7541  \\
        \bottomrule
    \end{tabular}
    \caption{MNIST validation results across priors and regularization methods. Values are averaged over
three seeds. The external PAk estimate has mean
$\bar{s}^{\mathrm{ext}} = -13.372 \pm 2.711$.}
\label{tab:mnist_result}
\end{table}

\section{Results}\label{Results}
We evaluate DiVAE on three synthetic datasets under both a fixed
standard-normal prior and a learnable GMM prior, using several external density
estimators (details in Appendix~\ref{Datasets} and~\ref{sec:exp-setup}). For
each configuration we compare a non-regularized baseline (\textsc{none}) to the
proposed alignment variants (\textsc{direct}, \textsc{flow}). We report:
(i) the mean prior log-density $\bar{s}$; (ii) distributional alignment between
model and external or ground-truth log densities using the two-sample
Kolmogorov--Smirnov (KS) test~\cite{massey1951} and the 1D Wasserstein distance (W)~\cite{villani2008optimal};
(iii) coverage between the true mixture and the learned prior via
$\mathrm{KL}(\text{true}\,\|\,\text{prior})$; and
(iv) the mean stochastic ELBO $\overline{\mathcal{L}}$.
For notation clarity, we summarize all quantities and definitions in
Tab.~\ref{tab:results-notation}. The reasoning behind our selection of metrics is discussed in Section \ref{sec:metrics_explanation}.

We also evaluate DiVAE on the MNIST dataset~\cite{lecun1998gradient}, where ground-truth log-densities
are not available. External densities are therefore estimated with the PAk
method~\cite{PAk2018}. We again compare \textsc{none}, \textsc{direct}, and \textsc{flow}
across three priors: standard normal, learnable GMM, and VampPrior. For each
model we report: (i) the mean prior log-density $\bar{s}$, interpreted as
latent log-likelihood; (ii) the ELBO $\overline{\ell}$; and (iii)
alignment with the PAk log-densities via KS and W.

Table~\ref{tab:ks-kl-main} summarizes results on the synthetic datasets, which consist of high-dimensional Gaussian mixtures with $k{=}4,8,12$ components (Appendix~\ref{Datasets}).
\textsc{direct} consistently improves KS and coverage KL under a learnable GMM
prior compared to \textsc{none}, whereas \textsc{flow} achieves the lowest KL
but with substantially higher KS, indicating over-correction and shape mismatch.
Under a fixed standard prior, \textsc{direct} generally improves KS, while
\textsc{flow} is beneficial primarily for more complex mixtures ($k{=}12$).
Complete numerical results are provided in
Tables~\ref{tab:results logs synthetic 2031965 oracle}--\ref{tab:results logs synthetic 14082005 PAk}.

Table~\ref{tab:mnist_result} reports MNIST results. Across all priors,
\textsc{direct} improves alignment with external PAk densities (lower KS and W)
while preserving or slightly improving the ELBO and $\bar{s}$ relative to
\textsc{none}. The \textsc{flow} variant often deteriorates both KS and W,
indicating that its stronger transformations can misalign the latent structure
in the absence of ground truth. Overall, \textsc{direct} provides the most
stable improvements across both synthetic and real data regimes.

\begin{table}[t]
    \centering
    \begin{tabular}{llcccc}
    \toprule
    Prior & Method &  $\bar{s}$$\downarrow$  & $KL(q,p_2)$$\uparrow$ & $\mathcal{H}(q)$$\uparrow$ \\
    \midrule
    GMM &  \textsc{none}            & -2.378  &  12.975 &  -5.785 \\
    GMM & \textsc{direct}      &  -3.365 &  12.477 & -5.216 \\
    GMM & \textsc{flow}        & \textbf{-7.397}  &  \textbf{16.567}  &  \textbf{-2.654} \\
    \midrule
    STANDARD &  \textsc{none}       &  -3.428 &   18.247  & -5.474 \\
    STANDARD & \textsc{direct} & -3.236  & 18.581 &  -5.645\\
    STANDARD & \textsc{flow}   &  \textbf{-4.850}  & \textbf{19.683} & \textbf{-5.326} \\
    \midrule
    \bottomrule
    \end{tabular}
    \caption{Out-of-distribution experiment on synthetic data. Models are trained on
a $k{=}4$ GMM and evaluated on an unseen $k{=}8$ GMM (dim=50). External densities for alignment
are estimated using PAk.}
    \label{tab:PAk-ood8}
\end{table}

\begin{table}[t]
    \centering
    \begin{tabular}{llcccc}
        \toprule
        Prior  & Method & $\Delta \overline \ell \downarrow$ & $\Delta \overline s \downarrow$ & $\Delta$KL $\uparrow$ & $\Delta \mathcal{H} \uparrow$\\
        \midrule
        STANDARD  & \textsc{none}   &  \textbf{-1058.67}  &  0.504  & \textbf{3.165}  &  -2.810  \\
        STANDARD  & \textsc{direct} &  -1025.35  &  \textbf{0.016}  & 1.554  &  \textbf{-1.577}  \\
        STANDARD  & \textsc{flow}   &  -1033.58  &  0.402  & 2.556  &  -3.169  \\
        \midrule
        
        GMM  & \textsc{none}   &  -1073.48  & -2.761 & 2.945  &  \textbf{0.450}  \\
        GMM  & \textsc{direct} &  -996.18   & \textbf{-3.608} & 3.476  &  0.091  \\
        GMM  & \textsc{flow}   &  \textbf{-1133.04}  & 0.400  & \textbf{3.615}  &  -4.202  \\
        \midrule
        
        VAMP & \textsc{none}   & -1063.15 & \textbf{-3.065}  & 2.055 &  0.725 \\
        VAMP & \textsc{direct} & -1071.64 & -2.790  & 0.916 &  \textbf{1.686}\\
        VAMP & \textsc{flow}   & \textbf{-1239.54} & 0.730  & \textbf{10.138} & -9.706  \\
        \bottomrule
    \end{tabular}
    \caption{OOD results on MNIST. Models are trained on MNIST and evaluated on
FashionMNIST. External densities for alignment are estimated using PAk.}
\label{tab:mnist_ood}
\end{table}

\paragraph{Out-of-distribution (OOD)}
We first assess OOD sensitivity on synthetic data by treating a $k{=}8$ GMM as
OOD for models trained on $k{=}4$ (Table~\ref{tab:PAk-ood8}). We report:
(i) the mean prior log-density $\bar{s}$, which decreases when the model
assigns lower latent likelihood to OOD samples; (ii) the posterior–true prior
divergence $\mathrm{KL}(q,p_2)$, capturing latent mismatch under OOD inputs; and
(iii) the posterior entropy $\mathcal{H}(q)$, reflecting uncertainty. 
Both \textsc{direct} and \textsc{flow} improve OOD separation over \textsc{none}
(lower $\bar{s}$). \textsc{flow} yields the strongest separation but also shows
substantially larger $\mathrm{KL}(q,p_2)$ and higher entropy, indicating
aggressive rejection and over-correction. \textsc{direct} provides a more
balanced trade-off: improved separation with moderate increases in divergence
and competitive entropy.

We further evaluate OOD detection on real data by training models on MNIST and
testing on FashionMNIST~\cite{xiao2017fashion} (Table~\ref{tab:mnist_ood}). Here we report shifts
between OOD and in-distribution metrics: ELBO ($\Delta \overline{\ell}$), prior log-density ($\Delta \bar{s}$), latent
prior mismatch shift ($\Delta\mathrm{KL}$), and posterior entropy shift ($\Delta\mathcal{H}$). Across all priors, \textsc{direct} yields
more stable and calibrated OOD shifts, with moderate ELBO and $\bar{s}$
reductions and controlled increases in $\Delta\mathrm{KL}$ and
$\Delta\mathcal{H}$. In contrast, \textsc{flow} produces the largest separation
(e.g.\ most negative $\Delta\overline{\ell}$ and largest $\Delta\mathrm{KL}$ for
GMM and VampPrior) but often at the cost of excessive mismatch and inflated
entropy. Overall, \textsc{direct} consistently improves OOD robustness especially on learnable priors, without
the over-rejection effects seen in the stronger \textsc{flow} transformation.

\begin{figure}
    \centering
    \includegraphics[width=1.0\linewidth]{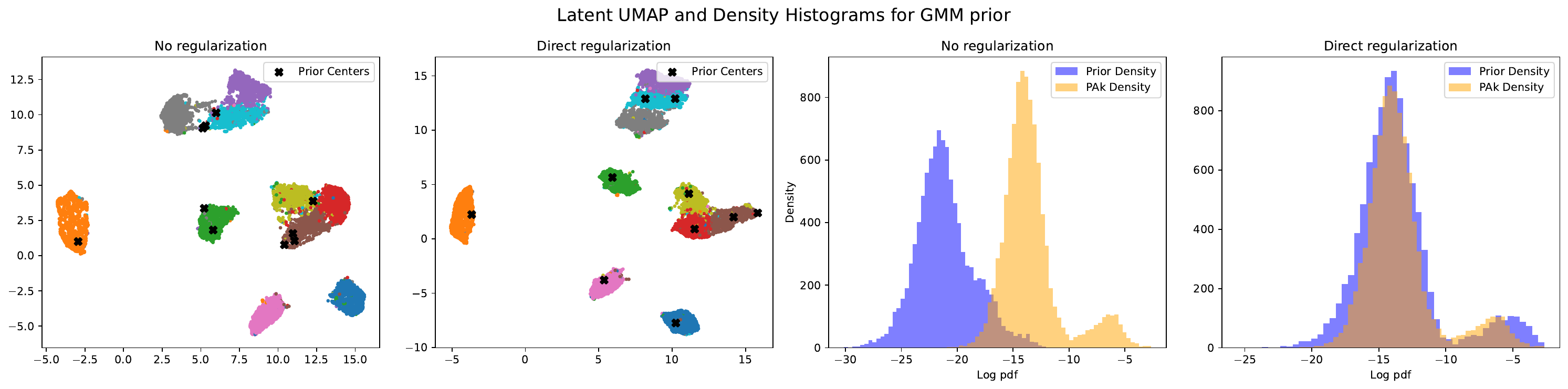}
    \caption{
UMAP representation and density alignments on the MNIST latent space. 
The regularized model places the prior components directly onto the data clusters 
(as evidenced by the coloring by ground-truth labels), whereas the unregularized 
model tends to collapse several prior centers onto the same region of the latent space. 
Moreover, the regularizer enforces a strong alignment between the learned log prior 
density and an external, non–black-box density proxy, providing a clear and 
interpretable correspondence between the latent geometry and the data distribution.
}
    \label{fig:mnist}
\end{figure}

\section{Conclusion}\label{Conclusions}
We introduced \emph{Density-Informed VAE (DiVAE)}, a lightweight, data-driven regularizer that aligns the VAE prior log-density with a log-density estimate derived from data via suitable aligners. On synthetic benchmarks with known ground truth, DiVAE improves prior log-density calibration under both standard-normal and GMM priors with negligible overhead. On MNIST, where the ground truth is unknown, we find similar benefits when we adopt a learnable prior (Figure~\ref{fig:mnist}).
These findings, although preliminary, suggest that explicitly coupling the prior to data-density structure can be a simple and effective route for developing new generative models or improving existing ones. Future directions include testing in real-world datasets and a deeper analysis of the promising flow approach.

\section*{Limitations}
The results we report are limited to simple datasets and small architectures: scaling to larger models and more challenging datasets remains to be demonstrated.

\newpage

\section*{Acknowledgements}

The authors acknowledge the AREA Science Park supercomputing platform ORFEO made available for conducting the research reported in this paper, and the technical support of the Laboratory of Data Engineering staff. Alessio Ansuini was supported by the project “Supporto alla diagnosi di malattie rare tramite l’intelligenza artificiale" CUP: F53C22001770002, from the project “Valutazione automatica delle immagini diagnostiche tramite l’intelligenza artificiale", CUP: F53C22001780002, and by the European Union – NextGenerationEU within the project PNRR “PRP@CERIC" IR0000028 - Mission 4 Component 2 Investment 3.1 Action 3.1.1.


\bibliographystyle{unsrt}
\bibliography{references}

\newpage

\appendix

\section{Datasets}\label{Datasets}
\section*{From 2D GMM to $D$-dimensional data via padding and rotation}
\paragraph{Setup.}
Let $u\in\mathbb{R}^2$ be drawn from a 2D mixture density $p_{2}(u)$ (the ground-truth Gaussian mixture in the main text).
For a target ambient dimension $D\ge 2$, set $m = D-2$ and define
\[
z \;=\; \begin{bmatrix} u \\ w \end{bmatrix}\in\mathbb{R}^D,
\qquad
w \sim \mathcal{N}\!\bigl(0,\,\sigma_{\text{pad}}^2 I_m\bigr).
\]
Let $R\in\mathbb{R}^{D\times D}$ be an orthogonal matrix ($R^\top R=I$) with $\det R=+1$, and set
\[
x \;=\; R z, \qquad z \;=\; R^\top x.
\]

By independence of $(u,w)$, the $D$-dimensional density factorizes as
\[
p_Z(z) \;=\; p_2(u)\,p_W(w)
\;=\; p_2(u)\,\prod_{j=1}^{m}\mathcal{N}\!\bigl(w_j;0,\sigma_{\text{pad}}^2\bigr).
\]

Since $x=Rz$ with orthogonal $R$, the change-of-variables determinant is unity ($|\det R|=1$), hence
\[
p_X(x) \;=\; p_Z\!\bigl(R^\top x\bigr).
\]
Writing $z=(u,w)=R^\top x$, we obtain the oracle log-density
\begin{equation}
\label{eq:true-log-pdf}
l^{\text{oracle}}(x) = \log p_X(x)
\;=\; \log p_2\!\bigl(u\bigr)
\;+\; \sum_{j=1}^{m}\log \mathcal{N}\!\bigl(w_j;0,\sigma_{\text{pad}}^2\bigr).
\end{equation}

\paragraph{Recovering the 2D generator density}
Let $R\in\mathbb{R}^{D\times D}$ be the rotation used to construct the dataset and define the selector
\[
S \;=\; \begin{bmatrix} I_2 & 0_{2\times(D-2)} \end{bmatrix}\in\mathbb{R}^{2\times D}.
\]
Given any observed point $x\in\mathbb{R}^D$, the unrotated coordinates are $z=R^\top x$, and the \emph{generator (ancestor) coordinates} in $\mathbb{R}^2$ are
\[
u \;=\; S\,z \;=\; S R^\top x.
\]
Equivalently, with the fixed linear map $\Pi = S R^\top \in \mathbb{R}^{2\times D}$,
\[
u \;=\; \Pi x, \qquad \Pi^\top \Pi \;=\; R S^\top S R^\top \;\text{ (projection onto the generator subspace)}.
\]
Since the global rotation is orthogonal, the 2D density of the latent ancestor corresponding to $x$ is simply the mixture density evaluated at $u$:
\begin{equation}
\label{eq:ancestor-density}
p_{\text{ancestor}}(x) \;=\; p_2\big(u\big) \;=\; p_2\!\big(\Pi x\big).
\end{equation}

\section{Experimental setup}\label{sec:exp-setup}
\paragraph{Synthetic data}
We construct three synthetic datasets as follows. We first sample from a 2D GMM with $k\in\{4,8,12\}$ mixture components, then append $(D{-}2)$ i.i.d.\ ``filler'' dimensions drawn from $\mathcal N(0,0.02^2)$. Subsequently, we apply a random orthogonal rotation in $\mathbb{R}^{D}$ as described in Appendix \ref{Datasets}.
We set $D{=}50$ for the first two datasets ($k{=}4$ and $k{=}8$) and $D{=}30$ for the third dataset ($k{=}12$). Each dataset contains $60{,}000$ training and $10{,}000$ validation points.

\paragraph{Priors}

We assess two priors for the VAE latent variable $z$: (i) the isotropic standard Gaussian $\mathcal{N}(0,I)$; and (ii) a learnable $k$-component Gaussian mixture,
$$
p_Z(z)=\sum_{k=1}^{k}\pi_k\,\mathcal{N}\!\big(z;\,\mu_k, \operatorname{diag}(\sigma_k^2)\big),
$$
with $k\in\{4,8,12\}$ chosen to match the generator’s component count. In the mixture prior, the component means $\{\mu_k\}$ and diagonal scales $\{\sigma_k\}$ are learned.

\paragraph{Density estimates}

We associate to each training sample its log-density estimate  $\rho_i$ used by the aligner as a supervising signal:
(i) \textit{Oracle}—the ground-truth $\log p_{X}(x)$ from the data generator (details on the Oracle are given in \ref{Datasets}, see in particular eq. \ref{eq:true-log-pdf});
(ii) \textit{PAk} estimator~\cite{PAk2018} computed after projecting the data onto the top $d{=}2$ principal components (matching the latent dimensionality). The PAk routine also returns a standard error, which we use as precision weight.

\paragraph{Training}
For the synthetic datasets, all models are VAEs with latent dimension $d = 2$, a Gaussian decoder with fixed observation noise $\sigma_x = 0.02$, and shallow linear MLPs. 
The encoder maps $D \rightarrow D/2 \rightarrow (\mu, \log \sigma^2) \in \mathbb{R}^2$, and the decoder maps $2 \rightarrow D/2 \rightarrow D$.
For MNIST, all models are VAEs with latent dimension $d = 10$, a Bernoulli decoder, and a shallow MLP with one hidden layer with $300$ neurons, both in the encoder and the decoder.

In both settings, we train with Adam ($lr=10^{-3}$) \cite{kingma2015adam}, batch size $128$, for $100$ epochs.
We apply a KL warm-up \cite{sonderby2016ladder} from $0.1$ to $1.0$ over the first half of training.
We compare \textsc{none} (no alignment), \textsc{direct} (direct aligner), and \textsc{flow} (flow aligner).
For the flow aligner, we use a small normalizing flow with coupling layers \cite{dinh2014nice,dinh2017realnvp,kingma2018glow}, with $5$ layers and hidden width $16$.
Experiments are repeated over multiple random seeds.\\
Our code is publicly available at \url{https://github.com/alessimichele/DiVAE-EurIPS}.

\section{Definitions and explanation of the metrics}\label{sec:metrics_explanation}

\textbf{Why KS/W.} We evaluate \emph{distributions} of log-densities through KS and W rather than relying on $\bar{s}$ alone, because means can mask “holes” (mass misplacement/missing modes): two models may match $\bar{s}$ yet allocate probability very differently. KS captures the worst-case gap between empirical CDFs, while W measures average transport. Together they provide complementary diagnostics of distribution-level calibration (lower is better).

\textbf{Why $\mathrm{KL}(p_2\,\|\,p_Z)$.} We favor KL(\emph{true} || \emph{prior}) because it penalizes under-allocation of probability where the \emph{true} generator places mass (poor coverage). Lower values indicate better coverage of the true support.

\textbf{Why $\mathrm{KL}(q\,\|\,p_2)$.} 
It quantifies how incompatible the latent encoding of $x$ is with the \emph{true} generator. 
For OOD inputs, a \emph{higher} value is desirable: it indicates the model correctly identifies that the posterior mass lies off the true latent manifold (i.e., it does not spuriously project OOD inputs onto plausible generator modes).

\textbf{Why entropy $\mathcal{H}(q)$.}
It measures the encoder’s uncertainty about the latent representation. 
OOD inputs should induce \emph{higher} entropy (broader posteriors), reflecting appropriate uncertainty rather than overconfident misassignments.

\textbf{Definition and motivation for $\Delta \mathrm{KL}$.} We define a \emph{latent mismatch shift} ($\Delta \mathrm{KL}$) as the difference between the expected posterior--prior divergence on OOD and in-distribution inputs:

\begin{equation}
    \Delta \mathrm{KL}
    \;=\;
    \mathbb{E}_{x \sim p_{\text{OOD}}}
        \big[ \mathrm{KL}\big(q(z \mid x)\,\|\,p(z)\big) \big]
    \;-\;
    \mathbb{E}_{x \sim p_{\text{IN}}}
        \big[ \mathrm{KL}\big(q(z \mid x)\,\|\,p(z)\big) \big].
\end{equation}

This metric measures how much does the posterior–prior mismatch increase when I move from in-distribution (MNIST) inputs to OOD (Fashion-MNIST) inputs, thus positive values of $\Delta \mathrm{KL}$ indicate that the posterior--prior 
mismatch is systematically larger on OOD data than on in-distribution data.

\textbf{Definition and motivation for $\Delta \mathcal{H}$.} We define an entropy-based \emph{uncertainty shift} as the difference between the expected posterior entropy on OOD and in-distribution inputs:
\begin{equation}
    \Delta \mathcal{H}
    \;=\;
    \mathbb{E}_{x \sim p_{\text{OOD}}}
        \big[ \mathcal{H}\big(q(z \mid x)\big) \big]
    \;-\;
    \mathbb{E}_{x \sim p_{\text{IN}}}
        \big[ \mathcal{H}\big(q(z \mid x)\big) \big],
\end{equation}
where $q(z \mid x)$ is the approximate posterior and $\mathcal{H}(\cdot)$ denotes
its entropy.

We hypothesize that, in a good model, the encoder becomes noticeably more uncertain when it sees OOD data. Therefore, positive values of $\Delta \mathcal{H}$ are to be preferred.

\section{Model details}\label{Model_details}

\paragraph{Huber loss}
The Huber loss, defined as:
\[
\Gamma_\delta(e)=
\begin{cases}
\frac{1}{2}e^2, & |e|\le \delta,\\[2pt]
\delta\!\left(|e|-\tfrac{1}{2}\delta\right), & |e|>\delta.
\end{cases}
\]

is quadratic when the residual’s magnitude $e$ is below a threshold $a$, and linear beyond that, with a continuous derivative at $a$. It is preferred to the squared error when performing robust regression for it is less sensitive to outliers \cite{huber1992robust}.

\section{Tables}
\begin{table}[h]
\centering
\resizebox{\textwidth}{!}{%
\begin{tabular}{l|l|l}
\toprule
\textbf{Notation} & \textbf{Definition} & \textbf{Notes} \\
\midrule
$\widehat{\mathcal{L}}_i$ 
  & $\log p_\theta(x_i\mid z_i)-\mathrm{KL}\!\big(q_\phi(z\mid x_i)\,\|\,p_Z(z;\psi)\big)$ 
  & Single-point ELBO estimator; $z_i \sim q_\phi(z\mid x_i)$ \\
\midrule
$\ell_i^{\mathrm{oracle}}$ 
  & $\log p_X(x_i)\quad(\text{see Appx.\ A.\ref{eq:true-log-pdf}})$ 
  & Single-point true data log-density \\
\midrule
$s_i$ 
  & $\log p_Z(z_i;\psi)$ 
  & Single-point log-prior probability; $z_i \sim q_\phi(z\mid x_i)$ \\
\midrule
$s_i^{\mathrm{oracle}}$ 
  & $\log p_2(u_i)\quad(\text{see Appx.\ A.\ref{eq:ancestor-density}}),\;\;u_i=\Pi x_i$ 
  & Single-point true generator (2D) log-density \\
\midrule
$\ell$ 
  & $(\widehat{\mathcal{L}}_i)_{i=1}^{N_{\text{val}}}$ 
  & Mean: $\overline{\ell}=\tfrac{1}{N_{\text{val}}}\sum_i \widehat{\mathcal{L}}_i$ \\
\midrule
$\ell^{\mathrm{oracle}}$ 
  & $(\ell_i^{\mathrm{oracle}})_{i=1}^{N_{\text{val}}}$ 
  & Mean: $\overline{\ell}^{\mathrm{oracle}}=\tfrac{1}{N_{\text{val}}}\sum_i \ell_i^{\mathrm{oracle}}$ \\
\midrule
$\mathbf{s}$ 
  & $(s_i)_{i=1}^{N_{\text{val}}}$ 
  & Mean: $\bar{s}=\tfrac{1}{N_{\text{val}}}\sum_i s_i$ \\
\midrule
$\mathbf{s}^{\mathrm{oracle}}$ 
  & $(s_i^{\mathrm{oracle}})_{i=1}^{N_{\text{val}}}$ 
  & Mean: $\bar{s}^{\mathrm{oracle}}=\tfrac{1}{N_{\text{val}}}\sum_i s_i^{\mathrm{oracle}}$ \\
\midrule
KS &  $\displaystyle \sup_x |F_{\mathbf{s}}(x)-F_{\mathbf{s}^{\mathrm{oracle}}}(x)|$    & Kolmogorov-Smirnov two samples test between $\bar{s}$ and $\bar{s}^{\mathrm{oracle}}$ \\
\midrule
W & $\displaystyle \int_0^1 \big|F_{\mathbf{s}}^{-1}(u)-F_{\mathbf{s}^{\mathrm{oracle}}}^{-1}(u)\big|\,du$ & 1D Wasserstein (EMD) between $\bar{s}$ and $\bar{s}^{\mathrm{oracle}}$ \\
\midrule
KL & $\mathrm{KL}(p_2(\cdot) \mid \mid p_Z(\cdot; \psi))$ & KL divergence between the true 2D GMM and the VAE's prior \\
\midrule
$KL(q,p_2)$ & $\mathrm{KL}(q(\cdot \mid x) \mid \mid p_2(\cdot) )$ & KL divergence between the variational posterior and the true 2D GMM\\
\midrule
$\mathcal{H}(q)$ & $-\mathbb{E}_{z\sim q_\phi(z\mid x_i)}\big[\log q_\phi(z\mid x_i)\big]$  & Entropy of the variational posterior\\
\bottomrule
\end{tabular}
}
\caption{Per-sample quantities and their vector/mean aggregates used in evaluation. $F$ denotes the empirical CDF.}
\label{tab:results-notation}
\end{table}

\begin{table}[h]
\centering
\scriptsize
\resizebox{\textwidth}{!}{%
\begin{tabular}{llcccccc}
\toprule
Prior & Method & $\overline{\ell}$& $\bar{s}$ & W & KS & KL \\
\midrule
GMM & No reg & \textbf{116.037 $\pm$ 5.193} & -1.998 $\pm$ 0.880 & 0.814 & 0.469 & 114.184  \\
GMM & \textsc{direct} & 115.970 $\pm$ 5.225 & \textbf{-2.589 $\pm$ 0.821} & \textbf{0.223} & \textbf{0.133} & 64.259 \\
GMM & \textsc{flow} & 115.856 $\pm$ 5.241 & -3.306 $\pm$ 0.806 & 0.516 & 0.398 & \textbf{34.723}  \\
\midrule
STANDARD & No reg & \textbf{115.408 $\pm$ 5.162} & \textbf{-2.728 $\pm$ 0.581} & \textbf{0.233} & \textbf{0.116} &   \\
STANDARD & \textsc{direct} & 115.225 $\pm$ 5.229 & -2.663 $\pm$ 0.528 & 0.258 & 0.126 &  \\
STANDARD & \textsc{flow} & 115.330 $\pm$ 5.219 & -2.964 $\pm$ 0.740 & 0.270 & 0.193 &  \\
\midrule
\bottomrule
\end{tabular}
}
\caption{Synthetic results for $k=4$, dim=$50$. Averaged across 3 seeds. Oracle external densities. \newline $\overline{\ell}^{\mathrm{oracle}}=$ 116.831; $\bar{s}^{\mathrm{oracle}}=$ -2.81252}
\label{tab:results logs synthetic 2031965 oracle}
\end{table}

\begin{table}[h]
\centering
\scriptsize
\resizebox{\textwidth}{!}{%
\begin{tabular}{llcccccc}
\toprule
Prior & Method & $\overline{\ell}$& $\bar{s}$ & W & KS & KL \\
\midrule
GMM & No reg & \textbf{116.037 $\pm$ 5.193} & -1.998 $\pm$ 0.880 & 0.814 & 0.469 & 114.184  \\
GMM & \textsc{direct} & 115.933 $\pm$ 5.186 & \textbf{-2.667 $\pm$ 0.759} & \textbf{0.182} & \textbf{0.102} & 69.620  \\
GMM & \textsc{flow} & 114.441 $\pm$ 5.291 & -6.395 $\pm$ 0.838 & 3.583 & 0.946 & \textbf{4.419}  \\
\midrule
STANDARD & No reg & \textbf{115.408 $\pm$ 5.162} & \textbf{-2.728 $\pm$ 0.581} & \textbf{0.233} & \textbf{0.116} &   \\
STANDARD & \textsc{direct} & 115.247 $\pm$ 5.182 & -2.473 $\pm$ 0.396 & 0.361 & 0.228 &   \\
STANDARD & \textsc{flow} & 115.082 $\pm$ 5.228 & -3.286 $\pm$ 0.813 & 0.500 & 0.338 &   \\
\midrule
\bottomrule
\end{tabular}
}
\caption{Synthetic results for $k=4$, dim=$50$. Averaged across 3 seeds. PAk external densities. \newline $\overline{\ell}^{\mathrm{oracle}}=$ 116.831; $\bar{s}^{\mathrm{oracle}}=$ -2.81252}
\label{tab:results logs synthetic 2031965 PAk}
\end{table}
\begin{table}[h]
\centering
\scriptsize
\resizebox{\textwidth}{!}{%
\begin{tabular}{llcccccc}
\toprule
Prior & Method & $\overline{\ell}$& $\bar{s}$ & W & KS & KL  \\
\midrule
GMM & No reg & 115.188 $\pm$ 5.300 & -2.495 $\pm$ 0.902 & 0.870 & 0.501 & 20.877 \\
GMM & \textsc{direct} & \textbf{115.245 $\pm$ 5.236} & -3.078 $\pm$ 0.813 & 0.294 & \textbf{0.187} & \textbf{8.840}  \\
GMM & \textsc{flow} & 114.755 $\pm$ 16.984 & \textbf{-3.117 $\pm$ 0.913} & \textbf{0.285} & 0.190 & 9.494  \\
\midrule
STANDARD & No reg & \textbf{114.339 $\pm$ 5.196} & -2.720 $\pm$ 0.496 & 0.645 & 0.322 &   \\
STANDARD & \textsc{direct} & 113.794 $\pm$ 17.043 & -2.847 $\pm$ 0.520 & 0.518 & 0.233 &   \\
STANDARD & \textsc{flow} & 113.748 $\pm$ 18.338 & \textbf{-3.054 $\pm$ 0.648} & \textbf{0.310} & \textbf{0.137} &   \\
\midrule
\bottomrule
\end{tabular}
}
\caption{Synthetic results for $k=8$, dim=$50$. Averaged across 3 seeds. Oracle external densities. \newline $\overline{\ell}^{\mathrm{oracle}}=$ 116.256; $\bar{s}^{\mathrm{oracle}}=$ -3.36459}
\label{tab:results logs synthetic 6032000 oracle}
\end{table}

\begin{table}[h]
\centering
\scriptsize
\resizebox{\textwidth}{!}{%
\begin{tabular}{llcccccc}
\toprule
Prior & Method & $\overline{\ell}$& $\bar{s}$& W & KS & KL  \\
\midrule
GMM & No reg & \textbf{115.188 $\pm$ 5.300} & -2.495 $\pm$ 0.902 & 0.870 & 0.501 & 20.877  \\
GMM & \textsc{direct} & 115.142 $\pm$ 5.305 & \textbf{-3.244 $\pm$ 0.803} & \textbf{0.151} & \textbf{0.073} & 6.447  \\
GMM & \textsc{flow} & 113.506 $\pm$ 5.891 & -6.325 $\pm$ 1.341 & 2.960 & 0.753 & \textbf{5.334}  \\
\midrule
STANDARD & No reg & \textbf{114.339 $\pm$ 5.196} & -2.720 $\pm$ 0.496 & 0.645 & 0.322 &  \\
STANDARD & \textsc{direct} & 113.919 $\pm$ 5.332 & -2.844 $\pm$ 0.504 & 0.521 & \textbf{0.231} &  \\
STANDARD & \textsc{flow} & 113.490 $\pm$ 17.099 & \textbf{-3.812 $\pm$ 0.925} & \textbf{0.493} & 0.315 &   \\
\midrule
\bottomrule
\end{tabular}
}
\caption{Synthetic results for $k=8$, dim=$50$. Averaged across 3 seeds. PAk external densities. \newline $\overline{\ell}^{\mathrm{oracle}}=$ 116.256; $\bar{s}^{\mathrm{oracle}}=$ -3.36459}
\label{tab:results logs synthetic 6032000 PAk}
\end{table}

\begin{table}[h]
\centering
\scriptsize
\resizebox{\textwidth}{!}{%
\begin{tabular}{llcccccc}
\toprule
Prior & Method & $\overline{\ell}$& $\bar{s}$ & W & KS & KL \\
\midrule
GMM & \textsc{none} & \textbf{64.961 $\pm$ 4.118} & -4.265 $\pm$ 0.925 & 0.633 & 0.380 & 3.728  \\
GMM & \textsc{direct} & 64.813 $\pm$ 4.141 & \textbf{-3.988 $\pm$ 0.773} & \textbf{0.334} & \textbf{0.238} & 3.349 \\
GMM & \textsc{flow} & 64.833 $\pm$ 4.155 & -5.049 $\pm$ 0.869 & 1.288 & 0.606 & \textbf{3.238}  \\
\midrule
STANDARD & \textsc{none} & \textbf{64.363 $\pm$ 4.069} & -2.866 $\pm$ 0.680 & 0.896 & 0.418 &  \\
STANDARD & \textsc{direct} & 64.156 $\pm$ 4.164 & -2.955 $\pm$ 0.723 & 0.808 & 0.388 & \\
STANDARD & \textsc{flow} & 64.272 $\pm$ 4.141 & \textbf{-3.199 $\pm$ 0.896} & \textbf{0.564} & \textbf{0.358} & \\
\midrule
\bottomrule
\end{tabular}
}
\caption{Synthetic results for $k=12$, dim=$30$. Averaged across 3 seeds. Oracle external densities. \newline $\overline{\ell}^{\mathrm{oracle}}=$ 66.045; $\bar{s}^{\mathrm{oracle}}=$ -3.76285}
\label{tab:results logs synthetic 14082005 oracle}
\end{table}

\begin{table}[h]
\centering
\scriptsize
\resizebox{\textwidth}{!}{%
\begin{tabular}{llcccccc}
\toprule
Prior & Method & $\overline{\ell}$& $\bar{s}$ & W & KS & KL \\
\midrule
GMM & \textsc{none} & \textbf{64.961 $\pm$ 4.118} & -4.265 $\pm$ 0.925 & 0.633 & 0.380 & \textbf{3.728}  \\
GMM & \textsc{direct} & 64.837 $\pm$ 4.051 & \textbf{-3.549 $\pm$ 0.608} & \textbf{0.239} & \textbf{0.124} & 6.823 \\
GMM & \textsc{flow} & 63.194 $\pm$ 4.300 & -8.195 $\pm$ 1.290 & 4.432 & 0.934 & 4.582  \\
\midrule
STANDARD & \textsc{none} & \textbf{64.363 $\pm$ 4.069} & -2.866 $\pm$ 0.680 & 0.896 & 0.418 &   \\
STANDARD & \textsc{direct} & 64.099 $\pm$ 4.173 & -2.996 $\pm$ 0.717 & 0.767 & 0.384 &  \\
STANDARD & \textsc{flow} & 63.967 $\pm$ 4.233 & \textbf{-3.824 $\pm$ 1.264} & \textbf{0.551} & \textbf{0.255} &  \\
\midrule
\bottomrule
\end{tabular}
}
\caption{Synthetic results for $k=12$, dim=$30$. Averaged across 3 seeds. PAk external densities. \newline $\overline{\ell}^{\mathrm{oracle}}=$ 66.045; $\bar{s}^{\mathrm{oracle}}=$ -3.76285}
\label{tab:results logs synthetic 14082005 PAk}
\end{table}

\begin{table}[t]
    \centering    
    \begin{tabular}{llccc}
        \toprule
        Prior & Method & $t_{\text{init}}$ & $\bar t_{\text{batch}}$ & $\bar t_{\text{epoch}}$ \\
        \midrule
        STANDARD & \textsc{none}   & - & 0.0040 & 1.898 \\
        STANDARD & \textsc{direct} & 12.08 & 0.0051 & 2.392 \\
        STANDARD & \textsc{flow}   & 10.68 & 0.0258 & 12.111 \\
        \midrule
        GMM      & \textsc{none}   & - & 0.0053 & 2.508 \\
        GMM      & \textsc{direct} & 12.34 & 0.0073 & 3.404 \\
        GMM      & \textsc{flow}   & 13.00 & 0.0295 & 13.849 \\
        \midrule
        VAMP      & \textsc{none}   & - & 0.0060 & 2.802 \\
        VAMP      & \textsc{direct} & 16.07 & 0.0088 & 4.133 \\
        VAMP      & \textsc{flow}   & 11.36 & 0.0309 & 14.482 \\
        \bottomrule
    \end{tabular}
    \caption{\textbf{Timing results.} 
    Direct regularization introduces essentially no computational overhead during training: 
    both per-batch and per-epoch times remain comparable to the unregularized baseline, with a 
    modest increase in initialization time due solely to the $O(N \log N)$ preprocessing step 
    required to compute the nearest-neighbour graph over the $N$ training samples. 
    In contrast, the flow-based regularizer incurs substantial overhead in both 
    batch and epoch times, as it requires training an additional normalizing flow module 
    jointly with the VAE.}
    \label{tab:mnist_timing}
\end{table}

\end{document}